\documentclass[10pt,twocolumn,letterpaper]{article}
\usepackage[pagenumbers]{cvpr}

\DeclareMathOperator*{\argmax}{arg\,max}

\usepackage{multirow}   
\usepackage{colortbl}   
\usepackage{adjustbox}  

\makeatletter
\newcommand{\teaserfig}[1]{\gdef\@teaser{#1}}
\let\@teaser\@empty
\g@addto@macro\@maketitle{\@teaser\par\vspace{8pt}}
\makeatother

\colorlet{grey}{gray}         
\colorlet{MidnightBlue}{MidnightBlue}
\newcommand{\model}{FluxMem}
\definecolor{isabelline}{HTML}{EAEAEA} 
\definecolor{lightblue}{HTML}{E6F1FF}  
\definecolor{deepgreen}{HTML}{006400}  
\definecolor{midgreen}{HTML}{2E8B57}   
\usepackage{pifont}
\usepackage{subcaption}
\newcommand{\cmark}{\textcolor{green!60!black}{\ding{51}}}
\newcommand{\xmark}{\textcolor{gray}{\ding{55}}}

\newcommand\blfootnote[1]{%
  \begingroup
  \renewcommand\thefootnote{}\footnote{#1}%
  \addtocounter{footnote}{-1}%
  \endgroup
}

\usepackage{amsmath}
\usepackage{xcolor}
\usepackage{algorithm}
\usepackage{algpseudocode}

\newcommand{\Encoder}{\mathrm{Encoder}}
\newcommand{\LLM}{\mathrm{LLM}}
\newcommand{\push}{\mathrm{push}}
\newcommand{\pop}{\mathrm{pop}}
\newcommand{\evictoldest}{\mathrm{evict\_oldest}}
\newcommand{\TAS}{\mathrm{TAS}}
\newcommand{\SDC}{\mathrm{SDC}}
\newcommand{\ACT}{\mathrm{ACT}}

\algtext*{EndIf}
\algtext*{EndWhile}
\algtext*{EndFor}
\algtext*{EndFunction}
\algtext*{EndProcedure}

\definecolor{exp_blue}{HTML}{2672B1}
\definecolor{exp_orange}{HTML}{E17227}
\usepackage{graphicx}
\usepackage{amssymb}

\definecolor{cvprblue}{rgb}{0.21,0.49,0.74}
\usepackage[pagebackref,breaklinks,colorlinks,allcolors=cvprblue]{hyperref}

\title{FluxMem: Adaptive Hierarchical Memory for Streaming Video Understanding}

\author{Yiweng Xie$^{1,2,3}$,~Bo He$^{4}$,~Junke Wang$^{1,3}$,~Xiangyu Zheng$^{1,3}$,~Ziyi Ye$^{1,3\dagger}$,~Zuxuan Wu$^{1,2,3\dagger}$
\\[0.5em]
{$^1$Institute of Trustworthy Embodied AI, Fudan University}, {$^2$Shanghai Innovation Institute} \\
{$^3$Shanghai Key Laboratory of Multimodal Embodied AI}, {$^4$University of Maryland, College Park}
\\[0.5em]
}

\teaserfig{
\vspace{-0.3in}
\centering
\includegraphics[width=0.95\linewidth,height=0.32\textheight,keepaspectratio]{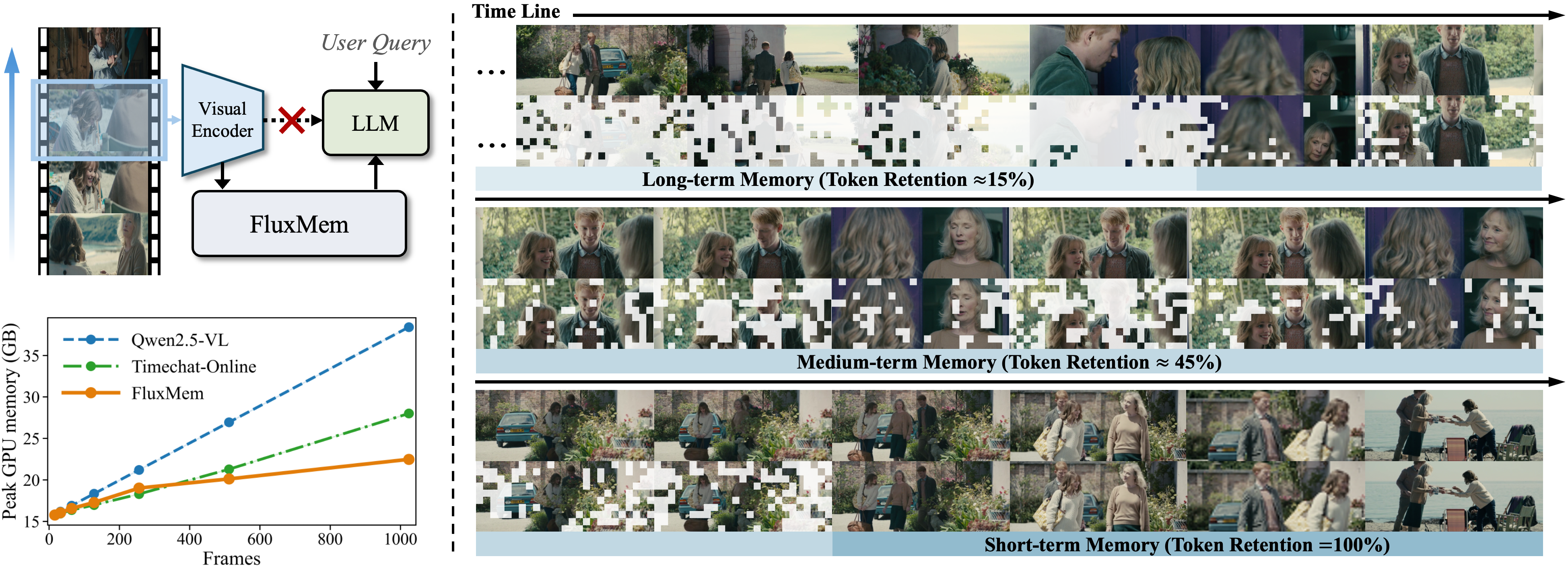}
\vspace{-0.1in}
\captionsetup{type=figure,hypcap=false,skip=+8pt}
\caption{{\model} is an adaptive hierarchical memory framework to progressively reduce visual tokens for efficient streaming video understanding. \textbf{Bottom left:} GPU memory usage vs. video frame length for {\model} and other methods during inference.}
\vspace{0.05in}
}

\begin{document}
\maketitle
\begin{abstract}
This paper presents \textbf{\model}, a training-free framework for efficient streaming video understanding. {\model} adaptively compresses redundant visual memory through a hierarchical, two-stage design: (1)~a Temporal Adjacency Selection~(TAS) module removes redundant visual tokens across adjacent frames, and (2)~a Spatial Domain Consolidation~(SDC) module further merges spatially repetitive regions within each frame into compact representations. To adapt effectively to dynamic scenes, we introduce a self-adaptive token compression mechanism in both TAS and SDC, which automatically determines the compression rate based on intrinsic scene statistics rather than manual tuning. Extensive experiments demonstrate that {\model} achieves new state-of-the-art results on existing online video benchmarks, reaching 76.4 on StreamingBench and 67.2 on OVO-Bench under real-time settings, while reducing latency by 69.9\% and peak GPU memory by 34.5\% on OVO-Bench. Furthermore, it maintains strong offline performance, achieving 73.1 on MLVU while using 65\% fewer visual tokens. Code is available at \href{https://github.com/YiwengXie/FluxMem}{https://github.com/YiwengXie/FluxMem}.
\blfootnote{$^{\dagger}$Corresponding author.}
\end{abstract}

\section{Introduction}
\label{sec:intro}
Recently, Multimodal Large Language Models (MLLMs) have achieved significant success in offline video understanding~\cite{openai2025gpt5,deepmind2025gemini,anthropic2025claude,bai2025qwen2_5vl,li2024llava_ov,wang2025internvl3_5,lei2021less,wang2022omnivl}.
However, real-world applications such as robotic manipulation~\cite{liu2025aligning}, autonomous driving~\cite{chen2023e2esurvey}, and smart glasses~\cite{lee2018interaction} demand real-time processing of streaming visual inputs.
This poses a critical challenge in effectively memorizing long-term temporal context and producing causal responses to user queries in an online manner~\cite{wang2023chatvideo,di2025rekv,ning2025livevlm,yang2025streammem,xu2025streamingvlm,zhang2025flash,yao2025timechat,xiong2025streamingvideounderstandingmultiround,wang2025streambridge,zeng2025streamforest}.

To process continuous visual streams in real-time, existing research has primarily focused on three directions: reducing visual tokens prior to the LLM~\cite{yao2025timechat,zeng2025streamforest}, managing the KV cache during the prefill stage~\cite{di2025rekv,ning2025livevlm,yang2025streammem},
and employing text query-guided filtering~\cite{ren2024timechat,yao2025timechat,chen2024image}.
Among these, token compression emerges as a more promising approach due to two advantages: (1)~unlike KV cache management, it performs deduplication before tokens enter the LLM, offering greater flexibility; (2)~unlike query-guided methods, it decouples textual and visual information, enabling the model to dynamically process the visual stream prior to query arrival. 
However, existing token compression methods such as TimeChat-Online~\cite{yao2025timechat} apply a single pruning/merging policy across the stream, overlooking the time-dependent utility of memory in streaming video.
Recent frames require dense preservation for grounding, while distant history tolerates stronger compression.
A global policy thus tends to under-prune long-term context and over-prune short-term details critical for causal reasoning.

To address this gap, we introduce \textbf{\model}, a training-free framework with an efficient adaptive visual memory design for streaming video understanding.
{\model} segments the visual context into three levels of memory: short-term, mid-term, and long-term, based on their temporal adjacency to the query moment. 
Accordingly, we propose a progressive visual token reduction strategy based on the hierarchical memory structure:
1)~Short-term Memory: all visual information is retained to preserve the immediate perceptual grounding for the current query;
2)~Mid-term Memory: Temporal Adjacency Selection ($\TAS$) is applied to reduce temporal redundancy by comparing adjacent frames and removing spatially aligned tokens;
3)~Long-term Memory: spatial redundancy within each frame is further reduced by grouping neighboring regions into representative anchors and removing repetitive tokens via Spatial Domain Consolidation ($\SDC$).
Unlike previous approaches~\cite{bolya2022tome,xiong2025streamingvideounderstandingmultiround,song2025moviechat+,yao2025timechat} that rely on manually tuned ratios for token retention or similarity-based thresholds, we derive frame-wise adaptive thresholds based on Otsu's method~\cite{otsu1979threshold} for token reduction in the mid- and long-term memory.
This adaptive design eliminates manual hyperparameter tuning and ensures robust performance across diverse scene dynamics.

We validate the effectiveness of {\model} on five video understanding benchmarks across both online and offline tasks.
The results demonstrate that {\model} achieves new state-of-the-art or highly competitive performance across diverse settings. For example, it achieves 53.3 on OVO-Bench~\cite{niu2025ovo} and 76.4 on StreamingBench~\cite{lin2024streamingbench} for streaming tasks, and 65.3 on VideoMME~\cite{fu2025video}, 73.1 on MLVU~\cite{zhou2025mlvu}, and 61.1 on LongVideoBench~\cite{wu2024longvideobench} for offline evaluation.
These results suggest that a single training-free memory framework can consistently improve online and offline video understanding with high efficiency and robustness.

In summary, our main contributions are as follows:
\begin{itemize}
\item We introduce a novel training-free hierarchical memory with two lightweight adaptive modules, which equips MLLMs with coherent short- and long-term video modeling for both online and offline settings.
\item Our approach achieves state-of-the-art performance on various video tasks in both online and offline settings while discarding 60--70\% of visual tokens and reducing latency and GPU memory usage.
\item We demonstrate that an adaptive token reduction threshold, based on video-specific information density, outperforms fixed-rule methods. This adaptive capability is natively supported by $\TAS$ and $\SDC$. 
\end{itemize}

\section{Related Work}
\label{sec:related}

\subsection{Multimodal Large Language Models}
Recent advances in Multimodal Large Language Models (MLLMs)~\cite{li2024llava_ov,wang2023see} have broadened their application to video understanding. 
Typically, these models comprise a visual encoder for extracting frame-level representations, a modality projector to map visual features into the language space, and a Large Language Model (LLM) to generate contextual responses~\cite{damonlpsg2023videollama,Maaz2023VideoChatGPT,bai2025qwen2_5vl,li2024llava_ov,zhang2024llavavideo,tang2025video,wang2025internvideo}. 
While these models achieve strong results on standard video benchmarks, they are inherently designed for static, offline settings where the input is the pre-loaded full video rather than a continuous stream. 
As a result, they fail to adapt to dynamic, real-world scenarios where video frames are processed sequentially and require real-time, temporally coherent, or even proactive responses~\cite{lin2024streamingbench,niu2025ovo,huang2025online}.

\subsection{Streaming Video Understanding}
In contrast to its offline counterpart, streaming video understanding requires models to sequentially process incoming video and generate real-time responses as user queries arrive~\cite{lin2024streamingbench,niu2025ovo,huang2025online,qian2024videostreaming}. In real-world scenarios such as robotic manipulation~\cite{liu2025aligning} and autonomous driving~\cite{chen2023e2esurvey}, models must efficiently manage historical information while focusing on the present context to produce timely and accurate reactions. 
To address these challenges, recent research has proposed several distinct approaches~\cite{nie2024slowfocus,wang2024omnivid,di2025rekv,ning2025livevlm,yang2025streammem,xu2025streamingvlm,zhang2025flash,wang2025streambridge,ding2025streammind,dos2025video}.
One line of research concentrates on memory design, which optimizes the storage and retrieval of historical information to alleviate the latency and GPU memory bottlenecks induced by large volumes of visual tokens~\cite{yang2025streammem,ning2025livevlm,zhang2025flash}.
A second strategy develops efficient visual token representations, compressing redundant information to improve computational efficiency without compromising semantic fidelity~\cite{bolya2022tome,yao2025timechat}.
A third approach introduces additional lightweight MLLMs to enable conventional offline models to proactively respond in streaming settings, effectively transforming them into online models~\cite{jin2024chatunivi,wang2025streambridge}.

\subsection{Visual Token Reduction} 
Among the numerous visual tokens in MLLM-based video understanding, only a small fraction carries essential semantics, while large portions correspond to repetitive or static regions that contribute little to the overall meaning. 
This redundancy not only causes significant computational and memory overhead during inference but also complicates cross-modal alignment.
To address this issue, visual token reduction is a common solution for the inherent redundancy in videos and the imbalance between the visual and language modalities~\cite{wang2022efficient,cheng2025vilamp,fu2025framefusion,shu2025video,shao2025tokens}.
Recent studies on visual token reduction can be categorized into three primary directions. 
The first focuses on memory compression or temporal redundancy modeling to aggregate long-term information and prevent the accumulation of unnecessary historical tokens~\cite{yang2025streammem,ning2025livevlm,he2024malmm}. A second strategy is to design adaptive token merging or pruning strategies that dynamically adjust token counts across spatial and temporal dimensions~\cite{bolya2022tome,yang2025visionzip,huang2024prunevid,tao2025dycoke,li2024llamavid,weng2024longvlm}. A third approach introduces language-guided mechanisms that leverage user queries to select the most relevant visual content and efficiently condense visual information~\cite{song2025moviechat+,di2025rekv}.

\section{Method}
\label{sec:method}

\begin{figure*}[t]
  \centering
  \includegraphics[width=1\linewidth]{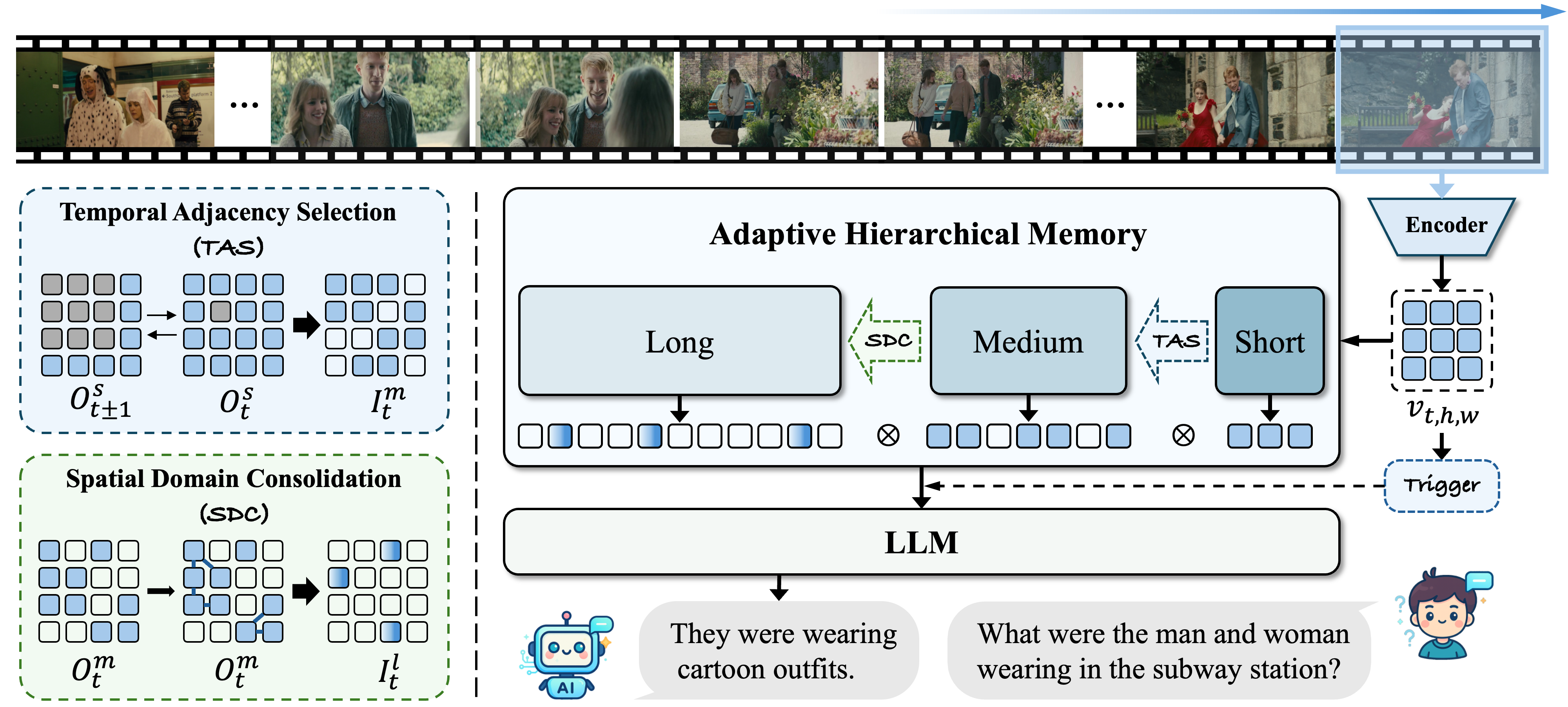}
  \caption{Overview of FluxMem: Adaptive Hierarchical Memory.
  Each incoming frame is encoded into visual tokens and written to {\model} in a cascaded short–mid–long process.
  On short-term memory overflow, Temporal Adjacency Selection ($\TAS$) retains temporally variant tokens for mid-term memory; on mid-term memory overflow, Spatial Domain Consolidation ($\SDC$) merges spatially redundant regions into compact anchors for long-term memory.
  The overflow process is guided by distribution-adaptive thresholds, autonomously calibrating retention strength to the video's temporal dynamics.
  Notably, the similarity metric against the preceding frame, required for $\TAS$, is computed upon the token's entry into the short-term memory, enabling it to serve as a zero-overhead trigger for active LLM output.}
  \label{fig:framework}
\end{figure*}

We propose \textbf{FluxMem} (Fig.~\ref{fig:framework}), an Adaptive Hierarchical Memory for spatiotemporal token reduction.
{\model} processes frames on the fly, retaining informative tokens and discarding redundancy in three cooperative parts.
(1) a streaming hierarchical memory with short-, mid-, and long-term memory (Sec.~\ref{subsec:shm}).
(2) two lightweight components, Temporal Adjacency Selection ($\TAS$) and Spatial Domain Consolidation ($\SDC$), to guide the dropping and compression of visual tokens (Sec.~\ref{subsec:reduction}).
(3) a distribution-adaptive thresholding scheme that derives data-driven thresholds from visual statistics (Sec.~\ref{subsec:Thresholding}).

\subsection{Streaming Hierarchical Memory}
\label{subsec:shm}
The streaming hierarchical memory maintains three cascaded modules to progressively compress while preserving causal consistency~\cite{atkinson1968human}. 
At timestep $t$, the streaming video frame $F_t$ is encoded by a vision encoder into a set of visual tokens $V_t \in \mathbb{R}^{HW \times D}$, where each token $v_{t,h,w} \in \mathbb{R}^{D}$ corresponds to spatial position $(h,w)$ within the frame. 
The memory is composed of $\mathcal{M}^s,\mathcal{M}^m,\mathcal{M}^l$ with capacity $c_s,c_m,c_l$. 
Visual tokens $V_t$ are first buffered in short-term memory. 
Tokens evicted from this level are temporally compressed by $\TAS$ (Sec.~\ref{par:TAS}) and stored in mid-term memory. If the token count exceeds capacity $c_m$, the earliest tokens are evicted to long-term memory, where they are spatially consolidated by $\SDC$ (Sec.~\ref{par:SDC}).
When a user query arrives or a proactive response is triggered (Sec.~\ref{par:proactive}), the tokens stored across the memory hierarchy are concatenated while preserving their spatiotemporal structure and then provided to the LLM as visual input for response generation.
The streaming procedure is summarized in Algorithm~\ref{alg:fluxmem}.

\subsection{Spatiotemporal Token Reduction}
\label{subsec:reduction}

To prune visual tokens, {\model} applies two lightweight, training-free modules: Temporal Adjacency Selection ($\TAS$) at the short-to-mid boundary and Spatial Domain Consolidation ($\SDC$) at the mid-to-long boundary as tokens progress forward through the memory hierarchy.
Concretely, we use the cosine distance $d(x, y) = 1 - \cos(x, y)$ and a per-frame, Otsu's method-derived threshold $\Theta_t$ (Sec.~\ref{subsec:Thresholding})~\cite{otsu1979threshold} that allows the policy to adapt automatically to scene dynamics.
With the arrival of new visual tokens, the hierarchy shifts from motion-centric selection ($\TAS$) to structure-centric consolidation ($\SDC$), turning short-lived dynamics into compact long-range context.

\begin{algorithm}[t]
\small
\caption{Streaming procedure of {\model}}
\label{alg:fluxmem}
\begin{algorithmic}[1]
  \Statex \textbf{Memory:} $\mathcal{M}^{s}, \mathcal{M}^{m}, \mathcal{M}^{l}$ with capacities $c_s, c_m, c_l$
  \Statex \textbf{modules:} $\TAS(\cdot), \SDC(\cdot), \ACT(\cdot)$; \textbf{Threshold:} $\gamma$

  \While{streaming frame $F_t$}
    \State $V_t \gets \Encoder(F_t)$;\; $r_t^{-} \gets \ACT(V_t)$;\; $\mathcal{M}^{s}.\push(V_t)$

    \State {\color{MidnightBlue}\textit{// response via query or activation trigger}}
    \If{(query $Q$ arrives) \textbf{or} $(t_a$ exists $\textbf{and}$ $ r_t^{-} > \gamma)$}
        \State $R \gets \LLM(\mathcal{M}^{l}\!\oplus\!\mathcal{M}^{m}\!\oplus\!\mathcal{M}^{s}, Q)$;\; $t_a \gets t$
    \EndIf

    \State {\color{MidnightBlue}\textit{// adaptive hierarchical memory update}}
    \If{$|\mathcal{M}^{s}| > c_s$}
        \State $O_t^{s} \gets \mathcal{M}^{s}.\pop;\; I_t^{m} \gets \TAS(O_t^{s});\; \mathcal{M}^{m}.\push(I_t^{m})$
    \EndIf
    \If{$|\mathcal{M}^{m}| > c_m$}
        \State $O_t^{m} \gets \mathcal{M}^{m}.\pop;\; I_t^{l} \gets \SDC(O_t^{m});\; \mathcal{M}^{l}.\push(I_t^{l})$
    \EndIf
    \If{$|\mathcal{M}^{l}| > c_l$}
        \State $\mathcal{M}^{l}.\evictoldest$
    \EndIf
    \State $t \gets t + 1$
  \EndWhile
\end{algorithmic}
\end{algorithm}

\paragraph{Temporal Adjacency Selection ($\TAS$).}
\label{par:TAS}
When the short-term memory overflows at time~$t$, $\TAS$ selectively retains tokens that exhibit significant semantic changes between adjacent temporal frames.
Since both neighboring frames are already stored in the short-term buffer when $\TAS$ is applied, the causal structure of the token stream is preserved.
For each spatial location $(h,w)$, we compute the token score as the minimum distance within a $3\!\times\!3$ window across adjacent frames:
\begin{equation}\label{eq:tas-minus}
s_{t,h,w}^{-} = \min_{(i,j)\in\mathcal{N}_{3\times3}(h,w)} d\big(v_{t,h,w}, v_{t-1,i,j}\big),
\end{equation}
\begin{equation}\label{eq:tas-plus}
s_{t,h,w}^{+} = \min_{(i,j)\in\mathcal{N}_{3\times3}(h,w)} d\big(v_{t,h,w}, v_{t+1,i,j}\big).
\end{equation}
Then we compute two separate, data-driven thresholds by applying Otsu's method (Sec.~\ref{subsec:Thresholding}) to the distributions of backward-looking scores ${s^{-}_{t,h,w}}$ and forward-looking scores ${s^{+}_{t,h,w}}$ independently, yielding $\Theta_t^{-}$ and $\Theta_t^{+}$, respectively.

A token $(h,w)$ is passed forward as the mid-term memory input if it is considered novel against \emph{either} the past or the future frame, i.e., if $(s^{-}_{t,h,w} > \Theta_t^{-}) \lor (s^{+}_{t,h,w} > \Theta_t^{+})$.
This union operation ensures that tokens are kept if they represent a significant change from the previous frame \emph{or} a significant change to the next frame.
Overall, by detecting redundant tokens via comparing each token to both adjacent frames with a small spatial window, $\TAS$ is robust to local misalignment from motion or camera jitter, while highlighting truly changing content without requiring optical flow or additional heads.
The procedure is strictly causal, single-pass, and $\mathcal{O}(HW)$ per overflow event.

\paragraph{Spatial Domain Consolidation ($\SDC$).}
\label{par:SDC}
When the mid-term memory reaches its capacity $c_m$, $\SDC$ consolidates locally redundant regions from an older frame by operating exclusively on the set of tokens already retained by $\TAS$.
For each retained token, we examine other retained tokens that lie within its original $3\!\times\!3$ spatial neighborhood and link them if their distance is $\leq\,\Theta_t$, which constructs a sparse, 8-connected graph defined only over this retained set.
Here, $\Theta_t$ is computed from the distribution of these pairwise spatial distances using Otsu's method (Sec.~\ref{subsec:Thresholding}).
A union-find pass yields connected components $\{\mathcal{C}_{t,k}\}_k$.
Each component is then replaced by its single mean anchor:
\begin{equation}\label{eq:sdc}
 a_{t,k} = \frac{1}{|\mathcal{C}_{t,k}|}\sum_{(i,j)\in\mathcal{C}_{t,k}} v_{t,i,j}.
\end{equation}

$\mathcal{A}_t = \{a_{t,k}\}_k$ is then appended to the long-term memory, with the union-find operator running in near-linear time complexity. 
Note that this graph is inherently sparse as it is built upon a pre-filtered set of tokens rather than the uncompressed tokens. Replacing each locally coherent region with its barycentric anchor, $\SDC$ removes the spatial redundancy while keeping necessary information.

\paragraph{Proactive Response Triggering.}
\label{par:proactive}
Online video understanding requires models to develop proactive output ability~\cite{wang2025streambridge,yao2025timechat,wang2024mmduet}. 
We implement this with a zero-cost trigger by reusing $\TAS$ statistics to detect scene changes and decide when to respond. Specifically, when frame $F_t$ enters short-term memory, $\TAS$ already computes the backward scores $s^{-}_{t,h,w}$ (Eq.~\eqref{eq:tas-minus}). Using these scores, we directly apply Otsu to $s^{-}_{t,h,w}$ to obtain $\Theta_t^{-}$ and define:
\begin{equation}\label{eq:trigger}
r_t^{-}=\tfrac{1}{HW}\sum_{h,w}\mathbf{1}\!\big[s^{-}_{t,h,w}>\Theta_t^{-}\big].
\end{equation}
We declare a scene switch when $r_t^{-}>\gamma$. Here, $\gamma\in[0,1]$ is a tunable parameter that controls the trigger sensitivity.

\begin{table*}[t]
  \centering
  \small
  \setlength{\tabcolsep}{4.5pt}
  \renewcommand{\arraystretch}{1.15}
  \caption{Results on the real-time subtasks of OVO-Bench and StreamingBench. \textbf{OVO-Bench real-time} encompasses: [OCR] Optical Character Recognition; [ACR] Action Recognition; [ATR] Attribute Recognition; [STU] Spatial Understanding; [FPD] Future Prediction; [OJR] Object Recognition. \textbf{StreamingBench real-time} encompasses: [OP] Object Perception; [CR] Causal Reasoning; [CS] Clip Summarization; [ATP] Attribute Perception; [EU] Event Understanding; [TR] Text-Rich Understanding; [PR] Prospective Reasoning; [SU] Spatial Understanding; [ACP] Action Perception; [CT] Counting. Best results among open-source models are in \textbf{bold} and the best results among training-free methods are \underline{underlined}. \small $^{\dagger}$ indicates the reproduced results.}
  \label{tab:results_real_time}
  \resizebox{\linewidth}{!}{
  \begin{tabular}{l c c @{\hspace{6pt}{\color{gray!60}\vrule width 0.3pt}\hspace{6pt}} c c c c c c @{\hspace{6pt}{\color{gray!60}\vrule width 0.3pt}\hspace{6pt}} c @{\hspace{6pt}{\color{gray!60}\vrule width 0.3pt}\hspace{6pt}} c c c c c c c c c c @{\hspace{6pt}{\color{gray!60}\vrule width 0.3pt}\hspace{6pt}} c @{\hspace{6pt}{\color{gray!60}\vrule width 0.3pt}\hspace{6pt}}}
    \toprule
    \multirow{2}{*}{Method} & \multirow{2}{*}{Size} & \multirow{2}{*}{Frames} & \multicolumn{7}{c}{OVO-Bench real-time} & \multicolumn{11}{c}{StreamingBench real-time} \\
    \cmidrule(lr){4-10} \cmidrule(lr){11-21}
    &&& OCR & ACR & ATR & STU & FPD & OJR & Avg. & OP & CR & CS & ATP & EU & TR & PR & SU & ACP & CT & Avg. \\
    \midrule

    \rowcolor{isabelline}
    \multicolumn{21}{c}{\textit{Human}} \\
    \midrule
    Human & -- & -- & 94.0 & 92.6 & 94.8 & 92.7 & 91.1 & 94.0 & 93.2 & 89.5 & 92.0 & 93.6 & 91.5 & 95.7 & 92.5 & 88.0 & 88.8 & 89.7 & 91.3 & 91.5 \\
    \midrule

    \rowcolor{isabelline}
    \multicolumn{21}{c}{\textit{Proprietary Models}} \\
    \midrule
    Gemini 1.5 Pro~\cite{team2024gemini} & -- & 1~fps & 85.9 & 67.0 & 79.3 & 58.4 & 63.4 & 62.0 & 69.3 & 79.0 & 80.5 & 83.5 & 79.7 & 80.0 & 84.7 & 77.8 & 64.2 & 72.0 & 48.7 & 75.7 \\
    GPT-4o~\cite{hurst2024gpt4o} & -- & 64 & 69.8 & 64.2 & 71.6 & 51.1 & 70.3 & 59.8 & 64.5 & 77.1 & 80.5 & 83.9 & 76.5 & 70.2 & 83.8 & 66.7 & 62.2 & 69.1 & 49.2 & 73.3 \\
    \midrule

    \rowcolor{isabelline}
    \multicolumn{21}{c}{\textit{Open-source Offline MLLMs}} \\
    \midrule

    LongVA~\cite{zhang2024longva} & 7B & 128 & -- & -- & -- & -- & -- & -- & -- & 70.0 & 63.3 & 61.2 & 70.9 & 62.7 & 59.5 & 61.1 & 53.7 & 54.7 & 34.7 & 60.0\\
    LongVU~\cite{shen2024longvu} & 7B & 1~fps & 55.7 & 49.5 & 59.5 & 48.3 & 68.3 & 63.0 & 57.4 & -- & -- & -- & -- & -- & -- & -- & -- & -- & -- & -- \\
    LLaVA-Video~\cite{zhang2024llavavideo} & 7B & 64 & 69.1 & 58.7 & 68.8 & 49.4 & 74.3 & 59.8 & 63.5 & -- & -- & -- & -- & -- & -- & -- & -- & -- & -- & -- \\
    LLaVA-OneVision~\cite{li2024llava_ov} & 7B & 64/32 & 66.4 & 57.8 & \textbf{73.3} & \textbf{53.4} & 71.3 & 62.0 & 64.0 & 80.4 & 74.2 & 76.0 & 80.7 & 72.7 & 71.7 & 67.6 & 65.5 & 65.7 & 45.1 & 71.1 \\
    InternVL2~\cite{chen2024internvl2} & 8B & 64/16 & 67.1 & \textbf{60.6} & 63.8 & 46.1 & 68.3 & 56.5 & 60.4 & 68.1 & 60.9 & 69.4 & 77.1 & 67.7 & 62.9 & 59.3 & 53.3 & 55.0 & 56.5 & 63.7 \\
    Qwen2.5-VL~\cite{bai2025qwen2_5vl} & 7B & 64 & -- & -- & -- & -- & -- & -- & -- & 78.3 & 80.5 & 79.8 & 82.4 & 75.5 & 80.4 & 74.1 & 62.6 & 67.6 & 51.1 & 73.9 \\
    \midrule

    \rowcolor{isabelline}
    \multicolumn{21}{c}{\textit{Open-source Online MLLMs (Training-Based)}} \\
    \midrule

    VideoLLM-Online~\cite{chen2024videollm}~\textsubscript{\textcolor{grey}{[CVPR 2024]}} & 8B & 2~fps & 8.1 & 23.9 & 12.1 & 14.0 & 45.5 & 21.2 & 20.8 & 39.1 & 40.1 & 34.5 & 31.1 & 46.0 & 32.4 & 31.5 & 34.2 & 42.5 & 27.9 & 36.0 \\
    Dispider~\cite{qian2025dispider}~\textsubscript{\textcolor{grey}{[CVPR 2025]}} & 7B & 1~fps & 57.7 & 49.5 & 62.1 & 44.9 & 61.4 & 51.6 & 54.6 & 74.9 & 75.5 & 74.1 & 73.1 & 74.4 & 59.9 & 76.1 & 62.9 & 62.2 & 45.8 & 67.6 \\
    Flash-VStream~\cite{zhang2025flash}~\textsubscript{\textcolor{grey}{[ICCV 2025]}} & 7B & 1~fps & 24.2 & 29.4 & 28.5 & 33.7 & 25.7 & 28.8 & 28.4 & 25.9 & 43.6 & 24.9 & 23.9 & 27.3 & 13.1 & 18.5 & 25.2 & 23.9 & 48.7 & 23.2 \\
    ViSpeak~\cite{fu2025vispeak}~\textsubscript{\textcolor{grey}{[ICCV 2025]}} & 7B & 1~fps & 75.2 & 58.7 & 71.6 & 51.1 & 74.3 & \textbf{66.9} & 66.3 & 79.8 & \textbf{88.3} & \textbf{83.3} & 81.1 & 76.4 & 75.1 & 70.4 & 65.9 & \textbf{77.3} & 34.2 & 74.4 \\
    TimeChat-Online~\cite{yao2025timechat}~\textsubscript{\textcolor{grey}{[ACM MM 2025]}} & 7B & 1~fps & 74.5 & 48.6 & 68.1 & 48.3 & 69.3 & 59.8 & 61.4 & 80.8 & 79.7 & 80.8 & 83.3 & 74.8 & 78.8 & 78.7 & 64.2 & 68.8 & \textbf{58.0} & 75.3 \\
    StreamForest~\cite{zeng2025streamforest}~\textsubscript{\textcolor{grey}{[NeurIPS 2025]}} & 7B & 1~fps & 68.5 & 53.2 & 71.6 & 47.8 & 65.4 & 60.9 & 61.2 & 83.1 & 82.8 & 82.7 & 84.3 & 77.5 & 78.2 & 76.9 & 69.1 & 75.6 & 54.4 & \textbf{77.3} \\

    \midrule
    \rowcolor{isabelline}
    \multicolumn{21}{c}{\textit{Open-source Online MLLMs (Training-Free)}} \\
    \midrule

     ReKV~\cite{di2025rekv}~\textsubscript{\textcolor{grey}{[ICLR 2025]}} & 7B & 0.5~fps & -- & -- & -- & -- & -- & -- & -- & 74.4 & 78.9 & 78.6 & 77.1 & 68.3 & 67.9 & 67.6 & 62.6 & 64.3 & 44.6 & 69.1 \\
     LiveVLM~\cite{ning2025livevlm}~\textsubscript{\textcolor{grey}{[arXiv 2025]}} & 7B & 0.5~fps & -- & -- & -- & -- & -- & -- & -- & \textbf{\underline{81.5}} & 78.1 & \textbf{\underline{83.3}} & 79.1 & 69.6 & 74.1 & 75.0 & \textbf{\underline{69.1}} & 67.7 & 40.4 & 72.9 \\

    \rowcolor{lightblue}
    Qwen2.5-VL$^{\dagger}$ & 7B & 1~fps & 79.2 & 53.2 & 67.2 & 51.7 & 71.3 & 57.1 & 63.3 & 78.3 & 80.5 & 79.8 & 82.4 & 75.5 & 80.4 & 74.1 & 62.6 & 67.6 & 51.1 & 73.9 \\

    \rowcolor{lightblue}
    \textbf{\model} & 7B & 1~fps 
    & \textbf{\underline{81.2}} 
    & \underline{59.6}
    & \underline{70.7}
    & \textbf{\underline{53.4}}
    & \textbf{\underline{75.2}}
    & \underline{63.0}
    & \textbf{\underline{67.2}}
    & 80.2
    & \underline{81.1}
    & 81.4
    & \textbf{\underline{85.3}}
    & \textbf{\underline{78.0}}
    & \textbf{\underline{83.8}}
    & \textbf{\underline{80.6}}
    & 65.9
    & \underline{69.6}
    & \underline{52.1}
    & \underline{76.4} \\
    \bottomrule
  \end{tabular}
  }
\end{table*}

\subsection{Adaptive Thresholding}
\label{subsec:Thresholding}

Both the $\TAS$ and $\SDC$ modules rely on a threshold to filter redundant tokens.
A fixed, manually-tuned threshold would fail to account for the highly variable dynamics of streaming video, performing poorly in scenes with either very low or high motion.
To ensure that our framework is robust and data-driven, we automatically compute all thresholds using Otsu's method~\cite{otsu1979threshold}. 
Otsu's method is a classic, non-parametric, and efficient algorithm that finds an optimal threshold to partition our similarity scores into two distinct groups for keeping and dropping.
It operates by exhaustively searching for the threshold $\theta$ that maximizes the inter-class variance $\sigma_B^2(\theta)$.
The optimal, data-driven threshold $\Theta_t$ at time $t$ is thus found by:

\begin{equation}\label{eq:otsu}
\Theta_t
=\argmax_{\theta}
\Bigl[
\omega_1(\theta)\omega_2(\theta)
\big(\mu_1(\theta)-\mu_2(\theta)\big)^2
\Bigr].
\end{equation}
where $\omega_1(\theta)$ and $\omega_2(\theta)$ represent the class probabilities, and $\mu_1(\theta)$ and $\mu_2(\theta)$ are the class means for the two clusters partitioned by the potential threshold $\theta$.
This process finds the partition that best separates the distribution.

Instead of using a fixed value, we compute the threshold $\Theta_t$ at runtime based on the target distribution: for $\TAS$, it analyzes the distribution of temporal similarity scores ($s^{-}_{t,h,w}$ and $s^{+}_{t,h,w}$) to adaptively determine the intensity of temporal change; while for $\SDC$, it takes into consideration the distribution of pairwise spatial distances to dynamically assess spatial adjacency relationships. This distribution-adaptive approach allows our reduction policies to automatically adjust their sensitivity to the input data, effectively raising the threshold in high-motion scenes and lowering the threshold in static ones without relying on learnable parameters or manual calibration.

\begin{table*}[t]
    \centering
    \footnotesize
    \setlength{\tabcolsep}{3pt}
    \renewcommand{\arraystretch}{1.05}
    \setlength{\aboverulesep}{0.2ex}
    \setlength{\belowrulesep}{0.2ex}

    \caption{Results on the online video understanding benchmarks \textbf{OVO-Bench} and \textbf{StreamingBench real-time}, as well as on the offline benchmarks \textbf{VideoMME (w/o sub.)}, \textbf{MLVU}, and \textbf{LongVideoBench}. Here, ``w/o sub.'' denotes evaluation on VideoMME without subtitles. Best overall results are in \textbf{bold} and the best results among training-free methods are \underline{underlined}. \small $^{\dagger}$ indicates the reproduced results.} 
    \label{tab:Benchmark_Others}
    \begin{adjustbox}{max width=\linewidth}
        \begin{tabular}{l c c @{\hspace{3pt}{\color{gray!30}\vrule width 0.2pt}\hspace{3pt}} c c c c c c c c}
            \toprule
            \multirow{3}{*}{Method} & \multirow{3}{*}{Size} & \multirow{3}{*}{Frames} & \multicolumn{2}{c}{Online Video} & \multicolumn{6}{c}{Offline Video} \\
            \cmidrule(lr){4-5} \cmidrule(lr){6-11}
            & & & StreamingBench & OVO-Bench & \multicolumn{4}{c}{VideoMME} & MLVU & LongVideoBench \\
            \arrayrulecolor{gray!30}\cmidrule(lr){6-9}\arrayrulecolor{black}
            & & & Real-Time & Overall & Short & Medium & Long & All & M-Avg & Val \\
            \midrule

            \rowcolor{isabelline}
            \multicolumn{11}{c}{\textit{Open-source Offline MLLMs}} \\
            \midrule

            LongVA~\cite{zhang2024longva} & 7B & 128 & 60.0 & -- & 61.1 & 50.4 & 46.2 & 52.6 & 56.3 & -- \\
            LongVU~\cite{shen2024longvu} & 7B & 1~fps & -- & 48.5 & -- & -- & -- & 60.6 & 65.4 & -- \\
            LLaVA-Video~\cite{zhang2024llavavideo} & 7B & 1~fps & -- & 53.1 & -- & -- & -- & 63.3 & 70.8 & -- \\
            LLaVA-OneVision~\cite{li2024llava_ov} & 7B & 32 & 71.1 & 52.9 & 70.1 & 56.4 & 48.8 & 58.4 & 64.7 & 56.5 \\
            InternVL2.5~\cite{chen2024internvl2_5} & 8B & 64 & -- & -- & -- & -- & -- & 64.2 & 68.9 & 60.0 \\
            Qwen2.5-VL~\cite{bai2025qwen2_5vl} & 7B & 768 & 73.7 & -- & -- & -- & -- & 65.1 & 70.2 & 56.0 \\
            \midrule

            \rowcolor{isabelline}
            \multicolumn{11}{c}{\textit{Open-source Online MLLMs (Training-Based)}} \\
            \midrule

            VideoLLM-Online~\cite{chen2024videollm}~\textsubscript{\textcolor{grey}{[CVPR 2024]}} & 8B & 2~fps & 36.0 & -- & -- & -- & -- & -- & -- & -- \\
            Dispider~\cite{qian2025dispider}~\textsubscript{\textcolor{grey}{[CVPR 2025]}} & 7B & 1~fps & 67.6 & 41.8 & -- & 53.7 & 49.7 & 57.2 & 61.7 & -- \\
            VideoChat-Online~\cite{huang2025online}~\textsubscript{\textcolor{grey}{[CVPR 2025]}} & 4B & 2~fps & -- & -- & -- & -- & 44.9 & 52.8 & -- & -- \\
            Flash-VStream~\cite{zhang2025flash}~\textsubscript{\textcolor{grey}{[ICCV 2025]}} & 7B & 1~fps & 23.2 & 33.6 & 72.0 & 61.1 & 50.3 & 61.2 & -- & -- \\
            TimeChat-Online~\cite{yao2025timechat}~\textsubscript{\textcolor{grey}{[ACM MM 2025]}} & 7B & 1~fps & 75.3 & 47.6 & -- & -- & 52.4 & 63.3 & 65.4 & 57.7 \\
            StreamForest~\cite{zeng2025streamforest}~\textsubscript{\textcolor{grey}{[NeurIPS 2025]}} & 7B & 1~fps & \textbf{77.3} & \textbf{55.6} & -- & -- & -- & 61.9 & 69.6 & -- \\
            
            \rowcolor{isabelline}
            \multicolumn{11}{c}{\textit{Open-source Online MLLMs (Training-Free)}} \\
            \midrule
            
            MovieChat~\cite{song2024moviechat}~\textsubscript{\textcolor{grey}{[CVPR 2024]}} & 7B & 2048 & -- & -- & -- & -- & 33.4 & 38.2 & -- & -- \\
            ReKV~\cite{di2025rekv}~\textsubscript{\textcolor{grey}{[ICLR 2025]}} & 7B & 0.5~fps & 69.1 & -- & -- & -- & -- & -- & 68.5 & -- \\
            StreamChat~\cite{xiong2025streamingvideounderstandingmultiround}~\textsubscript{\textcolor{grey}{[ICLR 2025]}} & 8B & 1~fps & 64.7 & -- & -- & -- & -- & -- & -- & -- \\
            LiveVLM~\cite{ning2025livevlm}~\textsubscript{\textcolor{grey}{[arXiv 2025]}} & 7B & 0.5/0.2~fps & 72.9 & -- & 66.7 & 56.4 & 48.8 & 57.3 & 66.3 & -- \\
            StreamMem~\cite{yang2025streammem}~\textsubscript{\textcolor{grey}{[arXiv 2025]}} & 7B & 4.0/0.5~fps & -- & -- & 71.5 & 62.4 & 52.3 & 62.1 & 65.9 & -- \\

            \rowcolor{lightblue}
            Qwen2.5-VL$^{\dagger}$ & 7B & 1~fps & 73.9 & 49.8 & 73.8 & 62.4 & 53.8 & 63.3 & 67.9 & 60.7 \\

            \rowcolor{lightblue}
            \textbf{\model} & 7B & 1~fps 
            & \underline{76.4} {\textcolor{midgreen}{(+2.5)}}
            & \underline{53.3} {\textcolor{midgreen}{(+3.5)}}
            & \textbf{\underline{76.9}} {\textcolor{midgreen}{(+3.1)}}
            & \textbf{\underline{65.1}} {\textcolor{midgreen}{(+2.7)}}
            & \textbf{\underline{54.0}} {\textcolor{midgreen}{(+0.2)}}
            & \textbf{\underline{65.3}} {\textcolor{midgreen}{(+2.0)}}
            & \textbf{\underline{73.1}} {\textcolor{midgreen}{(+5.2)}}
            & \textbf{\underline{61.1}} {\textcolor{midgreen}{(+0.4)}} \\

            \bottomrule
        \end{tabular}
    \end{adjustbox}
\end{table*}

\subsection{Discussion: Advantages of {\model}}
{\model} offers four key properties critical for streaming video understanding: (i) training-free operation, (ii) strict online causality, (iii) hierarchical memory organization, and (iv) adaptive compression.

\noindent\textbf{Training-free vs. Training-based.}
Online MLLMs, such as Flash-VStream~\cite{zhang2025flash}, require task-specific fine-tuning, which increases deployment costs and limits model generality. In contrast, {\model} addresses this limitation by operating as a training-free, plug-and-play module. It leverages on-the-fly similarity statistics through its $\TAS$ and $\SDC$ components, ensuring seamless compatibility with any pre-trained MLLM without the need for additional fine-tuning.

\noindent\textbf{Online-causal vs. Offline-global.}
Unlike conventional offline systems that rely on global keyframe selection~\cite{tang2025adaptive} or retrospective memory construction~\cite{shu2025video}, {\model} processes data in a strictly causal, hierarchical cascade. This design ensures real-time processing and long-horizon applicability, making {\model} well-suited for continuous streaming tasks where temporal causality must be maintained throughout.

\noindent\textbf{Hierarchical vs. Monolithic memory.}
Most prior memory designs rely on a flat buffer or a single-stage pruning module, which can disrupt temporal coherence or spatial structure. 
{\model} adopts a hierarchical, cascaded scheme: $\TAS$ preserves temporally novel tokens, while $\SDC$ consolidates spatial redundancies at a coarser granularity. 
This staged organization aligns with the intrinsic spatiotemporal hierarchy of video data, allowing the model to preserve salient long-range dependencies while yielding compact, spatial-preserving representations during streaming.

\noindent\textbf{Adaptive vs. Fixed Heuristics.}
Typical pruning-based methods~\cite{bolya2022tome,yao2025timechat} rely on fixed ratios or manually tuned thresholds, which fail to adapt to dynamic scene variations. In contrast, {\model} employs per-frame Otsu-based thresholds that automatically adjust the pruning strength, ensuring effective reduction while preserving important information in both static and fast-moving scenes.

\section{Experiments}
\label{sec:experiments}

\begin{figure*}[t]
  \centering
  \begin{subfigure}[t]{0.32\textwidth}
    \centering
    \includegraphics[width=\linewidth]{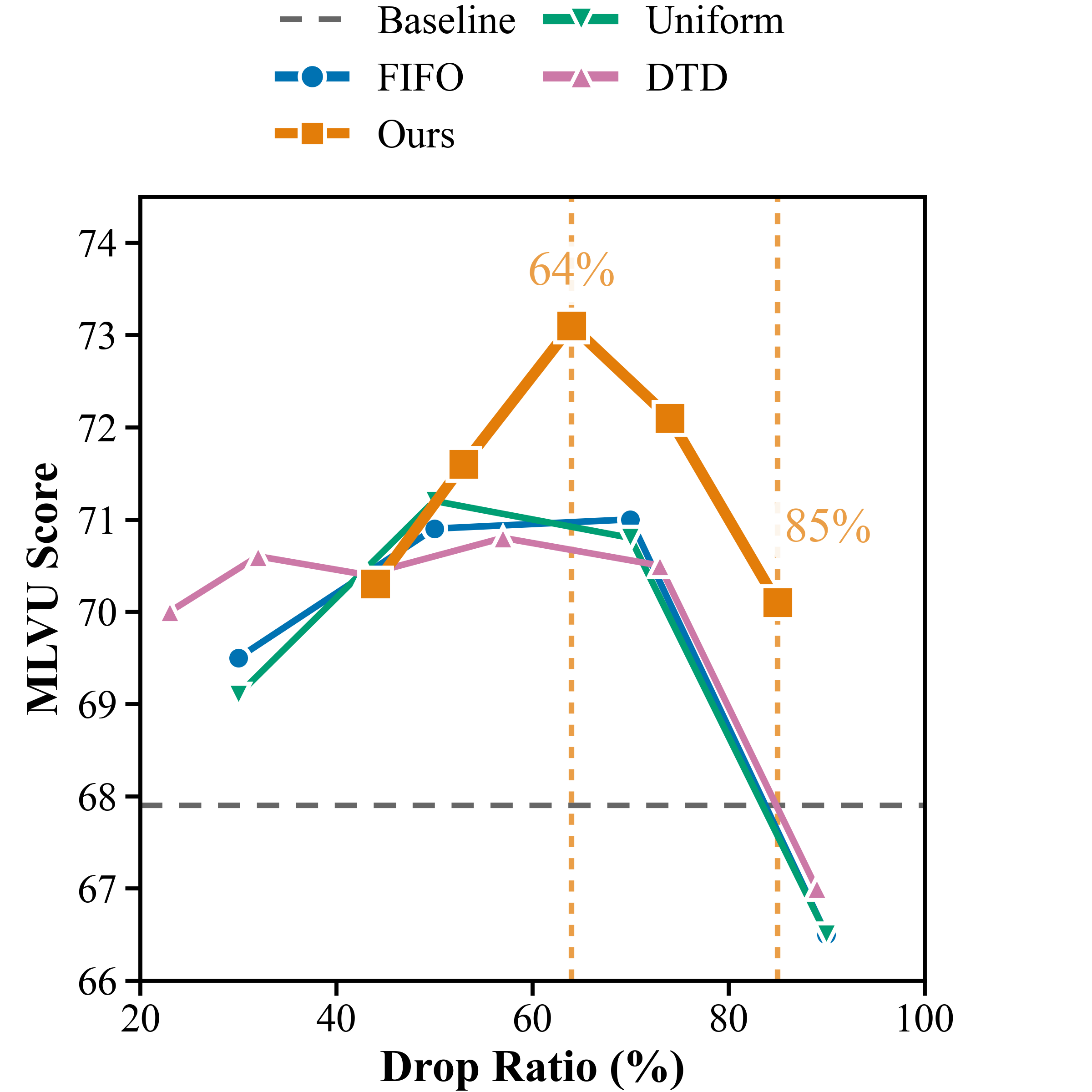}
    \caption{Memory bank compression methods ablation.}
    \label{fig:ablation_method}
  \end{subfigure}\hfill
  \begin{subfigure}[t]{0.32\textwidth}
    \centering
    \includegraphics[width=\linewidth]{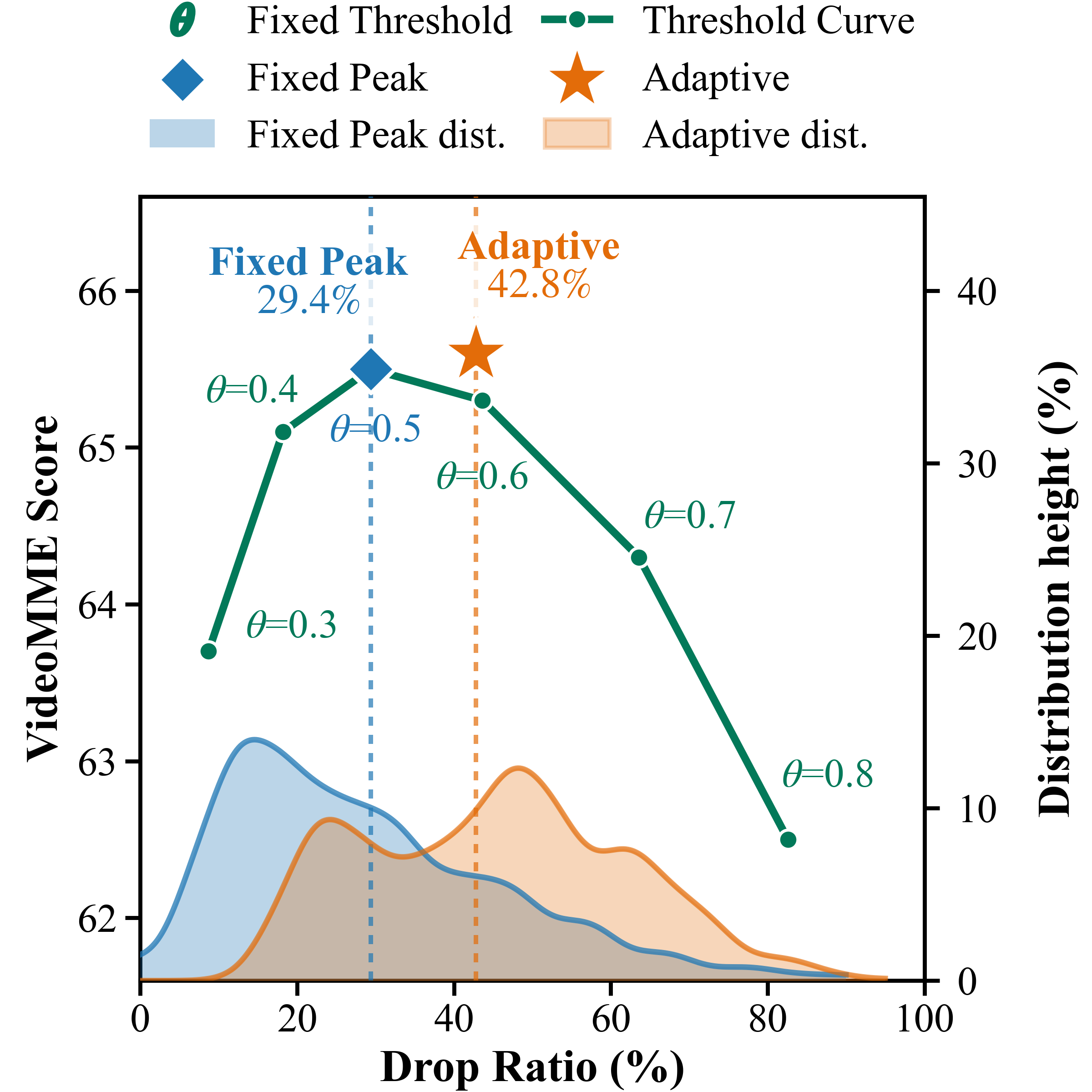}
    \caption{Mid-term memory drop curve.}
    \label{fig:ablation_mid}
  \end{subfigure}\hfill
  \begin{subfigure}[t]{0.32\textwidth}
    \centering
    \includegraphics[width=\linewidth]{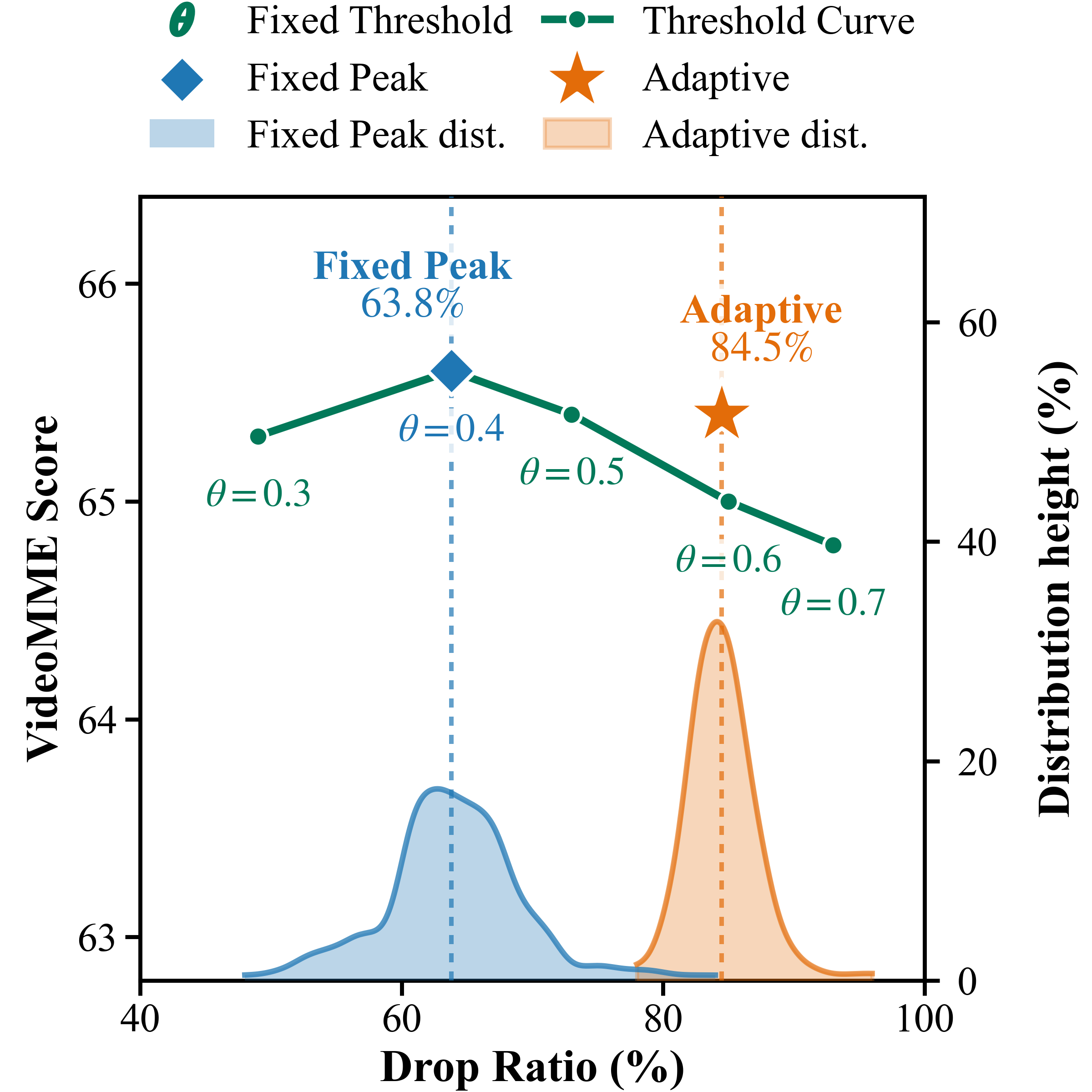}
    \caption{Long-term memory drop curve.}
    \label{fig:ablation_long}
  \end{subfigure}
  \caption{Ablation of {\model}. (a) Method comparison across drop ratios on the MLVU dataset. (b) and (c) Comparison of our adaptive and fixed thresholds in the mid- and long-term memory banks. The cosine distance of each token is compared against these thresholds to determine whether it is kept or dropped. The shaded area presents the distribution of average per-video drop ratios for the adaptive and optimal fixed thresholds, aggregated across all videos in the MLVU benchmark.}
  \label{fig:ablation_all}
\end{figure*}

\subsection{Implementation Details}
\vspace{-0.02in}
In our experiments, we implement {\model} based on Qwen2.5-VL-7B~\cite{bai2025qwen2_5vl}, since it is the current state-of-the-art video-language model.
For online benchmarks, videos are sampled at 1~fps to simulate real-time input, with up to 256 visual tokens per frame and 256 frames per video.
We set short-term memory length to 8 frames, mid-term memory length to 64, and assign the remaining frames to long-term memory, which can preserve long-range temporal context.
For offline benchmarks, we also use 1~fps, reducing per-frame tokens to 64 and limiting the maximum sequence length to 1024 frames.
We set short-, mid-, and long-term memory lengths as 8, 512, and the rest of the frames, respectively. 
This provides a balanced trade-off between efficiency and the global temporal coverage.
All experiments are conducted on 8~$\times$~A100 GPUs.

\subsection{Benchmarks}
\vspace{-0.02in}
To evaluate the effectiveness of the {\model}, we conduct experiments on both online and offline video understanding benchmarks.
For the online evaluation, we adopt OVO-Bench~\cite{niu2025ovo} and StreamingBench~\cite{lin2024streamingbench}.
OVO-Bench evaluates timestamp-based understanding of streaming video, covering historical retrieval, real-time awareness, and proactive response.  
StreamingBench evaluates models on real-time visual, omni-source, and contextual understanding. Our analysis concentrates on the real-time subtask. 
For the offline evaluation, we choose VideoMME~\cite{fu2025video}, MLVU~\cite{zhou2025mlvu}, and LongVideoBench~\cite{wu2024longvideobench}.
VideoMME offers full-spectrum multimodal assessment across diverse domains and durations; MLVU targets long-video multitask understanding with minutes- to hours-long clips; and LongVideoBench focuses on long-context referring reasoning with extended video-language sequences.

\subsection{Results on Online Video Understanding}
As shown in Table~\ref{tab:results_real_time}, {\model} achieves strong performance in real-time evaluation. On StreamingBench, {\model} improves the Qwen2.5-VL from 73.9 to 76.4 while compressing roughly 70\% of visual tokens. 
On OVO-Bench, {\model} improves real-time and overall from 63.3 to 67.2 and from 49.8 to 53.3, respectively~(Table~\ref{tab:Benchmark_Others}). 
Notably, significant gains could be observed in tasks that demand both short-horizon cues and stable temporal context, such as Prospective Reasoning~(+6.5) and Spatial Understanding~(+3.3) on StreamingBench, as well as Action Recognition~(+6.4) and Object Recognition~(+5.9) on OVO-Bench. 
We believe these improvements result from the hierarchical memory that preserves recent details while reducing temporal and spatial redundancy through $\TAS$ and $\SDC$ jointly. 
Among online MLLMs, {\model} achieves strong overall performance, surpassing LiveVLM~\cite{ning2025livevlm} and TimeChat-Online~\cite{yao2025timechat}. 
These results highlight the ability of {\model} to distill salient information from highly redundant video streams and deliver state-of-the-art online reasoning without task-specific training.

\subsection{Results on Offline Video Understanding}
As shown in Table~\ref{tab:Benchmark_Others}, {\model} outperforms the baseline on offline video QA benchmarks, despite being designed for online understanding and using significantly fewer tokens.
Specifically, it achieves 65.3~(vs.\ 63.3) on VideoMME, 73.1~(vs.\ 67.9) on MLVU, and 61.1~(vs.\ 60.7) on LongVideoBench.
These results surpass all training-free and training-based methods, validating {\model} as a highly generalizable, training-free framework that demonstrates strong efficacy on offline long-video benchmarks by excelling at salient information distillation.
Experiments on VideoMME subtasks validate the efficacy of {\model}. 
The model achieves significant gains in both short (from 73.8 to 76.9) and medium (from 62.4 to 65.1) contexts. 
More critically, performance in long contexts remains stable (from 53.8 to 54.0) even under an aggressive token reduction regime approaching 90\%.
These results indicate that the hierarchical design integrating $\TAS$ and $\SDC$ successfully retains salient tokens while discarding redundancy, thereby enhancing global understanding.

\subsection{Ablation Studies}
\label{subsec:ablation}

\paragraph{Efficiency.}
\begin{table}[t]
\centering
\footnotesize
\setlength{\tabcolsep}{4pt}
\caption{
Efficiency comparison between the baseline and FluxMem.
``$\downarrow$'' indicates reduction vs.\ baseline.
}
\label{tab:efficiency}
\begin{adjustbox}{max width=\linewidth}
\begin{tabular}{c l c c c}
\toprule
Dataset & Method & Latency$\downarrow$ (ms) & Peak Mem$\downarrow$ (GB) & Perf.$\uparrow$ \\
\midrule
\multirow{2}{*}{MLVU} 
     & Baseline & 3614 & 41.3 & 67.9 \\
     & FluxMem & 2014 ($\downarrow 44.3\%$) & 28.4 ($\downarrow 31.2\%$) & 73.1 \\
\midrule
\multirow{2}{*}{OVO-Bench} 
     & Baseline & 2701 & 35.8 & 49.8 \\
     & FluxMem & 812 ($\downarrow 69.9\%$) & 23.5 ($\downarrow 34.5\%$) & 53.3 \\
\bottomrule
\end{tabular}
\end{adjustbox}
\end{table}

Beyond accuracy, {\model} substantially improves deployment efficiency.
To isolate the contribution of the memory mechanism, we pre-extract frame-level visual features and measure the cost after vision encoding in Table~\ref{tab:efficiency}.
On OVO-Bench, {\model} cuts latency by 69.9\% and memory by 34.5\% while improving accuracy by +3.5.
On MLVU, it reduces latency and memory by 44.3\% and 31.2\% with +5.2 accuracy.
In the online setting, the per-frame update introduces only 4.1 ms overhead in total (1.3 ms for $\TAS$, 2.4 ms for $\SDC$ and 0.4 ms for others), preserving real-time efficiency.

\paragraph{Effect of Hierarchical Memory.} To evaluate the contributions of short-term (S), mid-term (M), and long-term (L) memory components, we conduct ablations by selectively enabling or removing each module. The results in Table~\ref{tab:ablation_memory_part} yield three consistent observations: (i) Complementary effects across different levels of memory. On MLVU, combining M and L reaches the highest accuracy of 73.1 while removing 65.6\% of visual tokens, outperforming models using only M or L. This indicates that $\TAS$ captures temporal variation and $\SDC$ captures spatial structure, and their roles become mutually reinforcing. (ii) Necessity of short-term memory for online perception. On real-time streaming tasks in StreamingBench, preserving the short-term memory stabilizes short-horizon perception. The S+L configuration attains an accuracy of 77.0, higher than using only S at 73.9 or only L at 75.9, showing that fine-grained short-range cues remain crucial for online understanding even when long-term memory is aggressively compressed. (iii) Balanced accuracy--efficiency trade-off. The full~(S+M+L) hierarchy attains the highest overall accuracy of 71.6 with 64.3\% of token reduction, remaining stable across both online and offline tasks. Using only S preserves too much redundancy, while using only L can remove subtle but important cues.

\begin{table}[t]
  \centering
  \footnotesize
  \setlength{\tabcolsep}{4pt}
  \renewcommand{\arraystretch}{1.1}
  \caption{Ablation of the memory module. ``$\downarrow$'' denotes average token drop ratio across all benchmarks. Best results are in \textbf{bold}.}
  \label{tab:ablation_memory_part}
  \begin{adjustbox}{max width=\linewidth}
  \begin{tabular}{c c c c @{\hspace{6pt}{\color{gray!60}\vrule width 0.3pt}\hspace{6pt}} c c c c}
    \toprule
    \multicolumn{4}{c}{Memory part} & MLVU & VideoMME & StreamingBench & Score \\
    \cmidrule(lr){1-4}
    S & M & L & Visual Tokens & M-Avg & w/o sub. & real-time & Avg. \\
    \midrule
    \cmark & \xmark & \xmark & $\downarrow 0.0\%$ & 67.8 & 63.3 & 73.9 & 68.3 \\
    \xmark & \cmark & \xmark & $\downarrow 43.2\%$ & 69.9 & 65.5 & 74.7 & 70.0 \\
    \xmark & \xmark & \cmark & $\downarrow 85.1\%$ & 70.9 & 62.0 & 75.9 & 69.6 \\
    \cmark & \cmark & \xmark & $\downarrow 42.4\%$ & 70.0 & \textbf{65.6} & 74.9 & 70.2 \\
    \cmark & \xmark & \cmark & $\downarrow 83.6\%$ & 70.9 & 61.9 & \textbf{77.0} & 69.9 \\
    \xmark & \cmark & \cmark & $\downarrow 65.6\%$ & \textbf{73.1} & 65.4 & 75.8 & 71.4 \\
    \rowcolor{lightblue}
    \cmark & \cmark & \cmark & $\downarrow 64.3\%$ & \textbf{73.1} & 65.3 & 76.4 & \textbf{71.6} \\
    \bottomrule
  \end{tabular}
  \end{adjustbox}
\end{table}

\paragraph{Comparison of Different Token Reduction Methods.}
Figure~\ref{fig:ablation_all}\subref{fig:ablation_method} compares our method with different token reduction strategies, including FIFO, Uniform, Random, and DTD~\cite{yao2025timechat}.
All methods exhibit a similar trend: performance first improves and then gradually declines as the drop ratio increases.
Within the practical 50--70\% range, {\model} consistently outperforms all counterparts, including the training-free DTD method used in TimeChat-Online~\cite{yao2025timechat}. Notably, {\model} achieves an accuracy of 73.1 on MLVU at a 64\% drop ratio, and still maintains 70.1 even under an aggressive 85\% token drop. We believe the superior performance stems from the structured hierarchical memory built upon the $\TAS$ and $\SDC$ modules.

\paragraph{Analysis of Adaptive Thresholding.}
We compare adaptive and fixed thresholds on MLVU~\cite{zhou2025mlvu} in Figure~\ref{fig:ablation_all}\subref{fig:ablation_mid} and Figure~\ref{fig:ablation_all}\subref{fig:ablation_long}. In the mid-term memory, the best fixed threshold achieves 65.5~(\textcolor{exp_blue}{\scalebox{1.2}[0.9]{$\blacklozenge$}} in \subref{fig:ablation_mid}) at a 29.4\% drop ratio, whereas the adaptive threshold attains 65.6~(\textcolor{exp_orange}{\scalebox{1}[1]{\ding{72}}} in \subref{fig:ablation_mid}) at a substantially higher 42.8\% drop ratio, indicating more efficient compression without sacrificing accuracy. In the long-term memory, the best fixed threshold reaches 65.6~(\textcolor{exp_blue}{\scalebox{1.2}[0.9]{$\blacklozenge$}} in \subref{fig:ablation_long}) at 63.8\% compression, while the adaptive threshold obtains a comparable 65.4~(\textcolor{exp_orange}{\scalebox{1}[1]{\ding{72}}} in \subref{fig:ablation_long}) at a much higher 84.5\% drop ratio. Beyond this accuracy--compression trade-off, the threshold distributions further reveal distinct behaviors across memory levels: in the mid-term memory, adaptive thresholds form a clear multi-peak distribution that reflects diverse inter-frame dynamics, whereas fixed thresholds collapse into a diffuse single mode; in the long-term memory, adaptive thresholds concentrate into a sharp single peak, suggesting a stable operating point induced by the more regular spatial redundancy and further reinforced by $\SDC$, while fixed thresholds remain scattered. Overall, these results show that adaptive thresholding flexibly captures temporal variability in mid-term memory while maintaining stable spatial reduction in long-term memory, consistently delivering stronger hierarchical token compression without manual tuning.

\paragraph{Training-free vs. SFT.}

\begin{table}[t]
  \centering
  \small
  \setlength{\tabcolsep}{4pt}
  \renewcommand{\arraystretch}{1.08}
  \caption{Ablation study of {\model} and SFT on the Qwen2.5-VL. Performance is evaluated on OVO-Bench (Overall) and StreamingBench (real-time). The SFT was performed on a subset of data from TimeChat-Online-139K~\cite{yao2025timechat} and VideoChatOnline-IT~\cite{huang2025online}.}
  \label{tab:ablation_sft}
  \begin{tabular}{lcc}
    \toprule
    Method & OVO-Bench & StreamingBench \\
    \midrule
    Qwen2.5-VL       & 49.8 & 73.9 \\
    + FluxMem        & 53.3\;\textcolor{midgreen}{(+3.5)} 
                     & 76.4\;\textcolor{midgreen}{(+2.5)} \\
    + FluxMem + SFT
                     & 61.4\;\textcolor{midgreen}{(+11.6)} 
                     & 76.7\;\textcolor{midgreen}{(+2.8)} \\
    \bottomrule
  \end{tabular}
\end{table}

As shown in Table~\ref{tab:ablation_sft}, {\model} not only improves the Qwen2.5-VL baseline under the training-free setting, but also remains competitive with supervised fine-tuning. Specifically, after applying SFT on a small subset of online video datasets~\cite{yao2025timechat,huang2025online}, the performance of our model is significantly improved on both OVO-Bench overall and StreamingBench real-time. 

\section{Conclusion}
\label{sec:conclusion}
This paper introduces {\model}, a training-free, plug-and-play framework that enables large multimodal models to effectively and efficiently process long streaming video sequences. {\model} maintains a hierarchical memory with two key components: Temporal Adjacency Selection and Spatial Domain Consolidation, which mitigate spatiotemporal redundancy while respecting causality. Experiments across both online and offline benchmarks have demonstrated that {\model} consistently outperforms all existing training-free methods, and even surpasses training-based methods. We believe {\model} offers valuable insights for future research on streaming video understanding. 
\section*{Acknowledgments}
This work was supported by the National Natural Science Foundation of China (No.62472098) and the Science and Technology Commission of Shanghai Municipality (No.24511103100).

{
    \small
    \bibliographystyle{ieeenat_fullname}
    \bibliography{main}
}


\end{document}